\documentclass{article}

\usepackage[nonatbib,final]{neurips_2021}

\usepackage[utf8]{inputenc} 
\usepackage[T1]{fontenc}    
\usepackage{hyperref}       
\usepackage{url}            
\usepackage{booktabs}       
\usepackage{amsfonts}       
\usepackage{nicefrac}       
\usepackage{microtype}      
\usepackage{xcolor}         
\usepackage{amsmath}
\usepackage{listings}
\usepackage[ruled]{algorithm2e}
\usepackage[frozencache,cachedir=.]{minted}
\usepackage{subcaption}
\usepackage{graphicx}

\usepackage[style=numeric, maxcitenames=1, uniquelist=false, uniquename=false, url =false, natbib]{biblatex}

\title{DARTS without a Validation Set: Optimizing the Marginal Likelihood}

\author{%
  Miroslav Fil$^1$ \thanks{Correspondence to miroslav.fil@cs.ox.ac.uk.}\\
  \And
  Binxin Ru$^1$ \\

  \And
  Clare Lyle$^1$ \\
  \And 
  Yarin Gal$^1$
  \AND
  \textsuperscript{1} \normalfont{OATML Group, Department of Computer Science, University of Oxford, UK}
}

\begin{document}

\maketitle

\begin{abstract}
  The success of neural architecture search (NAS) has historically been limited by excessive compute requirements. While modern weight-sharing NAS methods such as DARTS are able to finish the search in single-digit GPU days, extracting the final best architecture from the shared weights is notoriously unreliable. Training-Speed-Estimate (TSE), a recently developed generalization estimator with a Bayesian marginal likelihood interpretation, has previously been used in place of the validation loss for gradient-based optimization in DARTS. This prevents the DARTS skip connection collapse, which significantly improves performance on NASBench-201 and the original DARTS search space. We extend those results by applying various DARTS diagnostics and show several unusual behaviors arising from not using a validation set. Furthermore, our experiments yield concrete examples of the depth gap and topology selection in DARTS having a strongly negative impact on the search performance despite generally receiving limited attention in the literature compared to the operations selection. 
\end{abstract}

\section{Introduction}
Neural architecture search (NAS) algorithms have been able to automatically find network architectures that outperform the best human designs in test set performance on several benchmarks, making it an important subfield of AutoML \citep{elsken_neural_2019, ren_comprehensive_2021}. However, earliest specialized NAS methods using evolutionary algorithms \citep{real_regularized_2019} or reinforcement learning \citep{zoph_learning_2018} required thousands of GPU days to achieve good performance because each architecture was trained separately. Weight-sharing algorithms such as ENAS \citep{pham_efficient_2018} or DARTS \citep{liu_darts_2018}, in which all the architectures share the same set of weights in a large supernetwork, have reduced the computational cost to single-digit GPU days while often delivering better performance. However, identifying the single best architecture based on the shared weights is known to be unstable \citep{bender_understanding_2018, liu_darts_2018, zhang_how_2021}. DARTS in particular suffers from overfitting to too many skip connections \citep{zela_understanding_2019}, and \citet{dong_nas-bench-201_2020} have shown that DARTS might even select architectures with all skip connections and no parameter-based operations. 

Most weight-sharing NAS extracts the best architecture out of the supernetwork either by evaluating the validation accuracy of single architectures with weights inherited from the supernetwork \citep{li_random_2020, guo_single_2020}, or by explicitly optimizing the validation loss end-to-end in a differentiable fashion as in DARTS \citep{liu_darts_2018}. However, the performance of architectures using the shared weights tends to be significantly different from their performances when trained separately \citep{bender_understanding_2018, zhang_how_2021}, which makes it difficult to identify the standalone best architecture. 

\citet{ru_speedy_2021} have shown that Training-Speed-Estimate (TSE), a recent generalization estimator based on Bayesian model selection \citep{lyle_bayesian_2020} that does not require a validation set, can be used to provide high fidelity yet computationally cheap early-stopping estimates of architecture test set performance in non-weight-sharing NAS. They also showed promising results on TSE being applicable to various weight-sharing NAS algorithms. Our work extends those results by providing additional benchmarks of TSE applied to DARTS (TSE-DARTS) and its variants on the NASBench-series. The main novelty of our work lies in showing that using TSE both improves performance and fundamentally changes the behavior of DARTS, thus expanding the insights from previous work on explaining DARTS \citep{zela_understanding_2019,shu_understanding_2020}.

\section{Background}
We first introduce the basics behind DARTS and TSE, and then follow up by showing how to efficiently compute the gradients of TSE to use for the differentiable optimization in DARTS.

\subsection{DARTS}
DARTS \citep{liu_darts_2018} searches for a single architecture cell that is stacked repeatedly to form the final network. Such search space design is ubiquitous in modern NAS work \citep{pham_efficient_2018, real_regularized_2019, li_random_2020}. A cell in the DARTS search space is simply a directed acyclic graph (DAG) with 7 nodes. The DAG is fully connected, and every edge between a pair of nodes $(i, j)$ represents one operation $o$ from the 8 operations in the search space (such as skip connections, zero op, convolution or max pool). The output $x^j$ of intermediate nodes in the DAG is equal to a sum of the operations corresponding to all the edges $(\cdot, j)$ with input $x^i$:
\begin{equation}
    x^j = \sum_{i<j} o^{(i, j)} (x^i)
\end{equation}
During the search, DARTS relaxes the constraint that each edge $(i, j)$ must represent only a single operation, and instead uses a softmax with learnable weights $\alpha$ to make the sum differentiable. Therefore, the node output is a mixture of all the available operations at once defined as 
\begin{equation}
\label{eq:arch_softmax}
    \overline{o}^{(i,j)} = \sum_{o \in O} \frac{\exp (\alpha_o^{(i, j)})}{\sum_{o' \in O} \exp (\alpha_{o'}^{(i,j)})}o(x).
\end{equation}
DARTS searches for normal and reduction cells. Most of the network is composed of normal cells, whereas there are two reduction cells in total at $1/3$ and $2/3$ of the network depth. The difference is that reduction cells have all operations with stride two. The whole architecture encoding can thus be expressed as $\alpha = (\alpha_{normal}, \alpha_{reduce})$. DARTS formulates the optimization of the weights $w$ and architecture $\alpha$ as a bi-level optimization problem
\begin{subequations}
\begin{alignat}{2}
&\!\min_{\alpha}        &\qquad& L_{val}(w^*(\alpha), \alpha) \label{eq:outer}\\
&\text{s. t.} &      & w^*(\alpha) = \text{argmin}_w L_{train}(w, \alpha) \label{eq:inner}
\end{alignat}
\end{subequations}
$L_{train}$ and $L_{val}$ are losses on the train and validation sets, respectively. This formulation is equivalent to the training loops in other problems such as meta-learning \citep{finn_model-agnostic_2017} or gradient-based hyperparameter optimization \citep{baydin_automatic_2018}. The optimal weights $w^*$ are approximated by only a single step of SGD, which means that the architecture gradients actually descend $L_{val}(w-\eta \nabla_wL_{train}(w, \alpha))$. At the end of training, the supernetwork is discretized into a single architecture by taking argmax over each edge to select the operation that has the highest edge within the softmax. It is therefore implicitly assumed that the most useful operations will simultaneously have the highest weights. In practice, selecting the architecture this way is known to result in too many skip connections \citep{zela_understanding_2019, chen_progressive_2019}, which frequently results in poor performance of the final architecture.

\subsection{TSE}
TSE is defined as the sum over training losses during a model's optimization. Let $D = \{(x_1, y_1), (x_2, y_2), .., (x_n, y_n)\}$ be the training dataset, $f_{\theta_t}(x_{i(t)})$ represent a model's output for the training sample $x_{i(t)}$ in the $t$-th iteration with $\theta_t$ being the model weights at time $t$, $L$ be a loss function and $T$ be the total training iterations. We define TSE as:
\begin{equation}
\label{eq:TSE}
\text{TSE} = \sum_{t=1}^T L(f_{\theta_t}(x_{i(t)}), y_{i(t)}).
\end{equation}

\citet{lyle_bayesian_2020} have shown that TSE provably corresponds to an evidence lower bound in Bayesian linear regression, making it an estimate of the marginal likelihood. For non-linear models such as neural networks, TSE is a theoretically-inspired metric that was already shown to be useful as a predictor of generalization in NAS by \citet{ru_speedy_2021}. Similar optimization-based generalization estimators have also been noted to correlate well with generalization by \citet{jiang_fantastic_2020}. 

It is possible to integrate TSE into DARTS by replacing descent of the validation loss gradient $\nabla L_{val}(w^*(\alpha), \alpha)$ by descending the TSE gradient. Note that optimizing TSE amounts to minimizing the sum over training losses during training. Therefore, when using TSE-DARTS, no validation set is required, and we only use the training set to find both the weights and the architecture. We compute TSE over $T=100$ SGD iterations and iteratively update the architecture as in normal DARTS. Algorithm \ref{algo:TSE_darts} shows the TSE-DARTS training loop, which simply uses different gradients for updating the architecture compared to the original DARTS training loop \citep{liu_darts_2018}. We discuss how to compute the TSE gradient in Section \ref{sec:TSE_gradient}.

\begin{algorithm}[H]
    \label{algo:TSE_darts}
  Create mixed operations $\overline{o}^{(i,j)}$ parameterized by $\alpha^{(i,j)}$ for each edge $(i, j)$ \\
  Set T=100 steps of unrolling for computing TSE \\
  \While{not converged}{
    1. Approximate the optimal weights $w^*$ with T steps of SGD by computing $w_T = w_0 - \eta \sum_{t=0}^T \nabla_w L_{train}(f(x_t, w_t, \alpha), y_t)$\;
    2. Update architecture $\alpha$ by descending TSE gradient via Eq. (\ref{eq:foTSE_grad})\;
    3. Update the original weights with T steps of SGD $w_T = w_0 - \eta \sum_{t=0}^T \nabla_w L_{train}(f(x_t, w_t, \alpha), y_t)$ using the new architecture encoding;
  }
  \caption{TSE-DARTS}
\end{algorithm}

\subsection{Computing the TSE gradient}
\label{sec:TSE_gradient}
While optimizing TSE instead of the validation loss is conceptually simple, obtaining the TSE gradients is very computationally demanding in practice. Naively computing TSE gradient requires differentiating through the whole training history to compute gradients with respect to the architecture $\alpha$, which leads to $O(T)$ memory requirements due to the workings of reverse-mode auto-differentiation \citep{baydin_automatic_2018}. This would make it impossible to use large-scale networks. Precisely, the exact gradient for the final training loss at time $T$ after training with SGD is equal to (proof is included in Appendix \ref{app:TSE_grad}):
\begin{equation}
\label{eq:single_hypergradient}
    \nabla_{\alpha}L_{train}^T = \frac{\partial L_{train}^T}{\partial \alpha} + \frac{\partial L_{train}^T}{\partial w} (- \eta \sum_{0 \leq j \leq T}([\prod_{0 \leq k < j} I - \eta \frac{\partial^2 L_{train} ^{T-k-1}}{\partial w \partial w}] \frac{\partial^2L_{train}^{T-j-1}}{\partial w \partial \alpha})),
\end{equation}
where we abbreviate $L_{train}(f(x_T, w_T, \alpha), y_T)$ as $L_{train}^T$. The TSE gradient would be equivalent to summing such gradients over all the time steps $0, 1, .., T$. Instead, we follow \citet{ru_speedy_2021}, who proposed to use a first-order approximation for each training loss gradient that only includes the direct gradient $\frac{\partial L_{train}^T}{\partial \alpha}$ from Eq. (\ref{eq:single_hypergradient}) at each time step. The whole TSE gradient can then be approximated as
\begin{equation}
\label{eq:foTSE_grad}
    \nabla_{\alpha} \text{TSE} = \nabla_{\alpha} (L_{train}^0 + L_{train}^1 + .. + L_{train}^T) \approx \sum_{t=1}^T \frac{\partial L_{train}^t}{\partial \alpha}
\end{equation}
This form of the TSE gradient has several advantages. Most importantly, the $\nabla_{\alpha} L_{train}^t$ can be computed for free concurrently with the weights gradients. As a result, computing the approximate TSE gradient has the same time and memory costs as normal training via backprop. An example implementation is shown in Appendix \ref{app:TSE_darts_implementation}.

\section{Experiments}
We first show the performance of TSE-DARTS compared to baseline DARTS on NASBench-201 \citep{dong_nas-bench-201_2020} and NASBench-1shot1 \citep{zela_nas-bench-1shot1_2020}. While those search spaces are small, they tabulate the ground truth test set performance of every architecture in the search space. The tabular nature of those benchmarks makes it especially viable to use those spaces for diagnosing the behavior of TSE-DARTS, which we do in Sections \ref{sec:darts_shallow} and \ref{sec:darts_ev} as our main empirical contribution. For the DARTS search space, we retrain our best architectures using the 600 epochs training protocol as originally used by DARTS \citep{liu_darts_2018}. Moreover, we use the DARTS search protocol, where we first run the search four times, retrain all the resulting architectures once, and then retrain the best out of the four for three more seeds to compute its mean ground truth test set accuracy. We also evaluate TSE versions of DrNAS \citep{chen_drnas_2021} and PDARTS \citep{chen_progressive_2019}. Our experiments with TSE algorithms always reuse the default parameters from their corresponding public implementations. 

\subsection{NASBench-201 and NASBench-1shot1}
Searching on NASBench-201 \citep{dong_nas-bench-201_2020} is a notorious example of DARTS overfitting to skip connections as the final selected architecture often contains only skip connections with no parameter-based operations. This makes DARTS have a consistently poor performance. Table \ref{tab:nb201_101} shows that using TSE-DARTS is able to prevent the skip connection collapse and achieve $92.66\%$ test set accuracy, which is competitive with other weight-sharing algorithms such as GDAS \citep{dong_searching_2019} or SPOS \citep{li_random_2020} in NASBench-201. On the other hand, DARTS has a strong performance on all three search spaces from NASBench-1shot1 \citep{zela_nas-bench-1shot1_2020}, from which we show results on search space 3, denoted as NB101-3. Note that none of the search spaces have skip connections in the operations set, hence DARTS cannot overfit to them. In comparison, TSE-DARTS fares poorly on this benchmark. Section \ref{sec:darts_shallow} shows that this is due to picking cells that are too deep rather than due to the operations choice, which is unusual given the consistent overfitting to skip connections in normal DARTS. Appendix \ref{app:extra_nb201_nb101} includes results on additional NASBench-201 datasets and NASBench-1shot1 search spaces, which are mostly analogous to those already shown.
\begin{table}[t]
\centering
\begin{tabular}{@{}lll@{}}
\toprule
             &                                                          & CIFAR10                          \\ \midrule
NASBench-201 & SPOS        \citep{li_random_2020}    & 91.05 (0.66)                     \\
             & GDAS  \citep{dong_searching_2019}     & \multicolumn{1}{c}{93.23 (0.58)} \\
             & ENAS      \citep{pham_efficient_2018} & 93.76 (0.00)                     \\
             & DARTS \citep{liu_darts_2018}          & 65.38 (7.84)                     \\
             & TSE-DARTS                                               & 92.66 (0.00)                     \\ \midrule
NB101-3      & DARTS \citep{liu_darts_2018}                                                    & 93.35 (0.01)                     \\
             & TSE-DARTS                                               & 87.87 (0.00)                     \\ \bottomrule
\end{tabular}
\caption{Summary of the performance of TSE-DARTS on NASBench-201 and NASBench-1shot1 search space 3 (abbreviated as NB101-3). TSE-DARTS has a much stronger performance than DARTS on NASBench-201 as it is able to avoid collapsing to all skip connections. However, DARTS does better on NB101-3 because of better topology as discussed in Section \ref{sec:darts_shallow}.}
\label{tab:nb201_101}
\end{table}
\subsection{DARTS search space and the role of depth}
\label{sec:darts_depth}
Table \ref{tab:darts_orig_results} shows the results of TSE-DARTS on the original DARTS search space. The best architecture found by TSE-DARTS achieves $2.73\%$ test set accuracy, which is a marginal improvement over the $2.76\%$ of second-order DARTS. However, it is a significant improvement compared to the $3.00\%$ of first-order DARTS, which is particularly notable since both TSE-DARTS and first-order DARTS have the same compute and memory requirements. Moreover, TSE-DARTS finds very different architectures compared to normal DARTS. Even in the DARTS search space, there is no overfitting to skip connections for TSE-DARTS. Instead, the discovered architectures tend to be very heavy on separable convolutions, often with no parameter-free operations at all. The best architecture found by TSE-DARTS is visualized in Appendix \ref{app:best_arch}.

\begin{figure}[t]
\centering
\begin{subfigure}{.33\textwidth}
  \centering
  \includegraphics[scale=0.31]{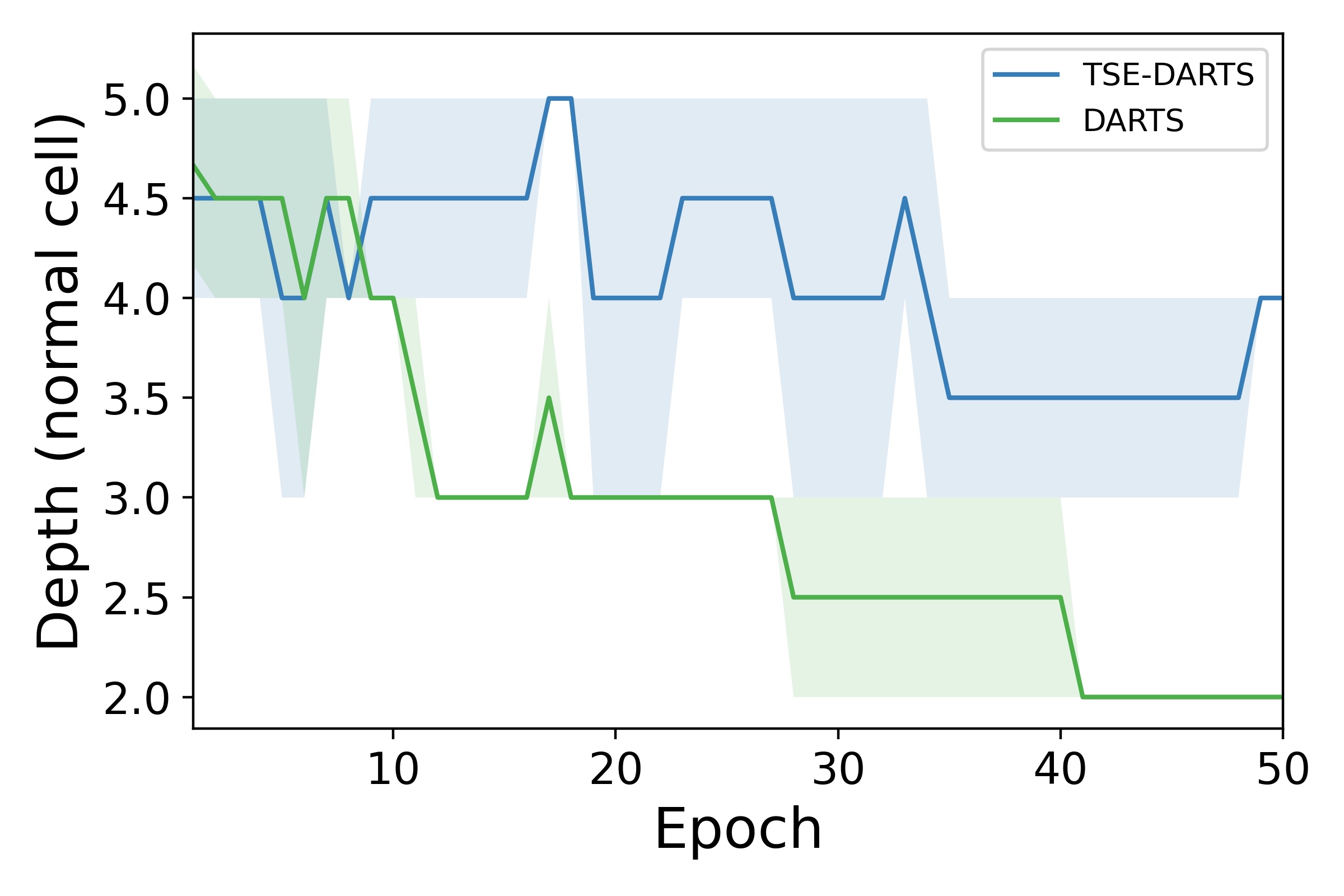}
    \caption{Normal cell (8 layers)}

  \label{fig:darts_depth_normal}
\end{subfigure}%
\begin{subfigure}{.33\textwidth}
  \centering
  \includegraphics[scale=0.31]{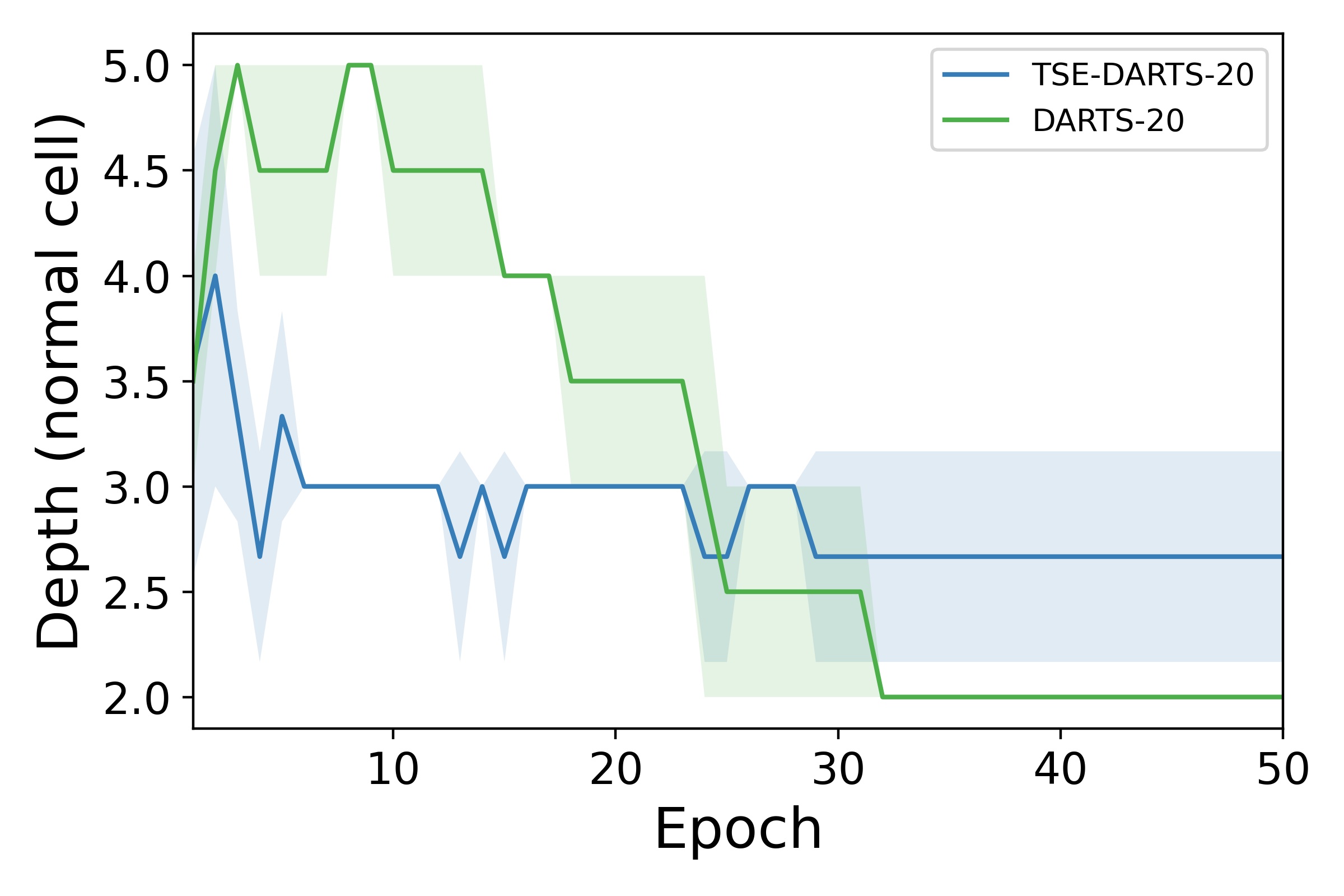}
  \caption{Normal cell (20 layers)}
  \label{fig:darts_depth_normal_20}
\end{subfigure}%
\begin{subfigure}{.33\textwidth}
  \centering
  \includegraphics[scale=0.31]{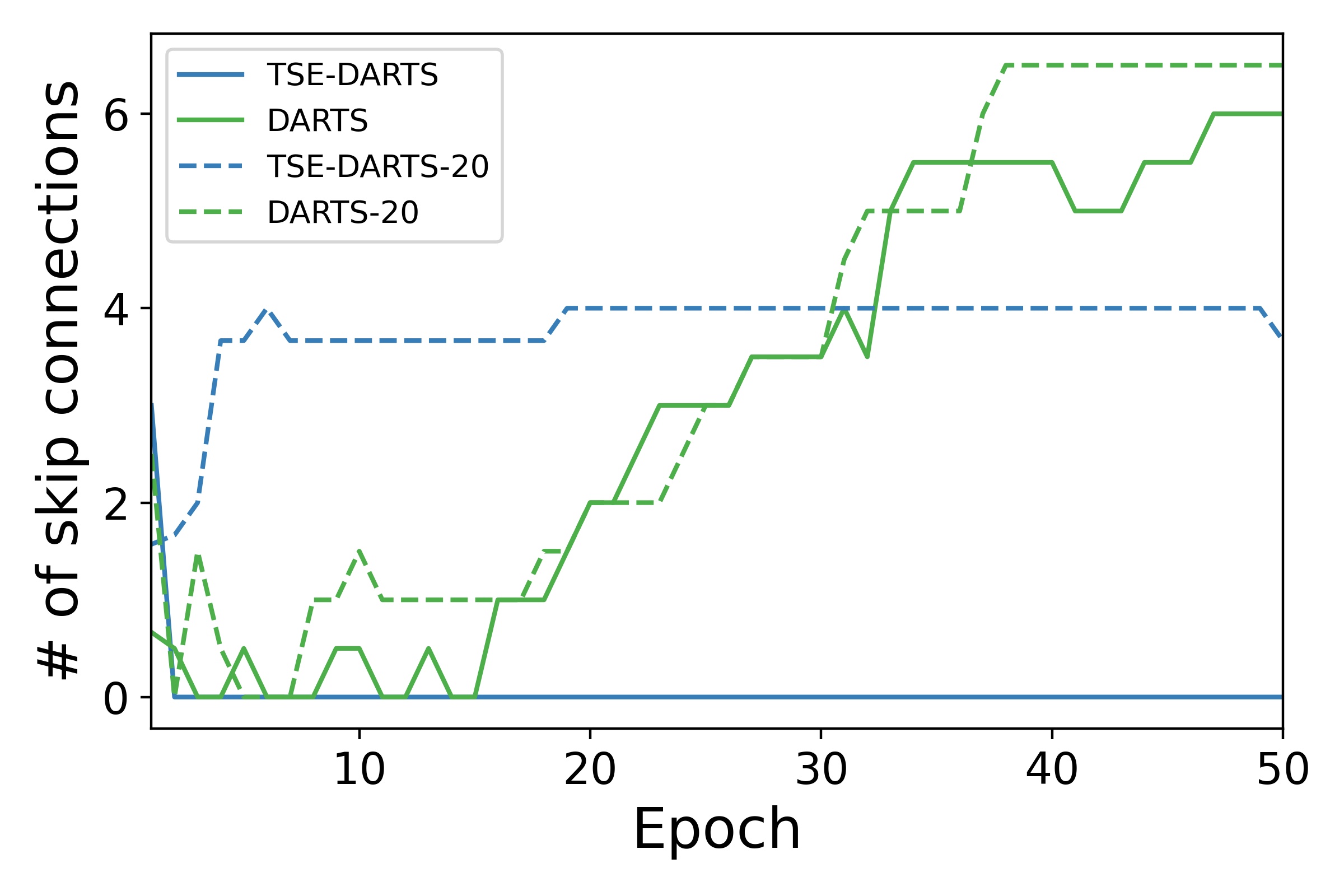}
  \caption{\# of skip}
  \label{fig:darts_skip}
\end{subfigure}
\caption{When searching on the DARTS search space, baseline DARTS is biased towards shallow architectures. Architectures found by TSE-DARTS tend to be significantly deeper for both 8 and 20 layers search as shown in a) and b), respectively. The range of depth in the DARTS space is [2, 5], hence DARTS attains the minimum depth possible in the normal cell. c) shows the number of skip connections, where we note that TSE-DARTS-20 picks a stable, moderate amount of skip connections while DARTS overfits to them over time.}
\label{fig:darts_depth}
\end{figure}

We hypothesize that skip connections are never chosen because they are unnecessary in the DARTS search setup, which uses a shallow search supernetwork. In contrast, note that skip connections are known to be more beneficial for deeper architectures \citep{he_deep_2016, li_visualizing_2018}. The search uses only 8 layers in the supernetwork even though the evaluation with retraining from scratch is done using 20 layers networks. This inequity between the search and evaluation networks is often referred to as the depth gap. To test whether TSE-DARTS would pick skip connections with a deeper search supernetwork, we also ran the search using 20 layers (denoted as TSE-DARTS-20) instead of the original 8 layers. Figure \ref{fig:darts_skip} compares the number of skip connections for TSE-DARTS and TSE-DARTS-20. The increased depth clearly favors picking a moderate amount of skip connections without overfitting to them as in original DARTS, whereas the shallow TSE-DARTS has no skip connections whatsoever. The best architecture of TSE-DARTS-20 has standalone test accuracy of $2.68\%$, further improving on the $2.73\%$ of TSE-DARTS. We also benchmarked first-order DARTS-20, which reached $2.89\%$ final test set accuracy, improving on the $3.00\%$ of original first-order DARTS.

To better assess the effects of depth gap, we also trained the standalone architectures from scratch using 8 layers for the evaluation. This again completely eliminates the depth gap except that now both the search and evaluation networks are shallow. Table \ref{tab:darts_orig_results} shows the results with 8 layers evaluation using the same architectures discovered by each search algorithm as the corresponding 20 layers evaluation. In this setup, the TSE variants consistently improve on their baseline versions. This is true particularly for TSE-DrNAS, which has a weak performance of $2.90\%$ in the original 20 layers evaluation setup, but top tier results in 8 layers evaluation. Because some of the TSE variants significantly improve in the 8 layers evaluation but are close or worse than the respective baselines using 20 layers evaluation, it suggests that the architectural optimum with 8 layers networks can be quite different from the 20 layers optimum. Hence, even if the search algorithm itself is improved, it might still lead to worse results in the DARTS 20 layers evaluation if the search only uses 8 layers, since the performance gains might not transfer across different depths due to the depth gap.

Related work including PDARTS \citep{chen_progressive_2019} and DrNAS \citep{chen_drnas_2021} tries to tackle the depth gap by progressively deepening the search network while removing some operations from the operations set during multiple phases of the search. We also tried to do this and tested TSE-PDARTS, but we found the resulting architectures to be very poor with only $3.19\%$ accuracy. Furthermore, the final architectures never had any skip connections because the skip connections already got removed from the search space while the network was still shallow. This suggests that even though the progressive deepening might be a key factor in PDARTS's impressive $2.50\%$ performance, the improvement might not necessarily be coming from reducing the depth gap since the architectures found by TSE-DARTS-20 and TSE-PDARTS are very different.

\begin{table}[t]
\centering
\begin{tabular}{@{}lcc@{}}
\toprule
                                                        & \begin{tabular}[c]{@{}c@{}}8 layers eval\\ (\% err. rate)\end{tabular} & \begin{tabular}[c]{@{}c@{}}20 layers eval\\ (\% err. rate)\end{tabular} \\ \midrule
DARTS (1st) \citep{liu_darts_2018}   & 4.87 (0.18)                                                            & 3.00 (0.14)                                                             \\
DARTS (2nd) \citep{liu_darts_2018}   & 4.91 (0.22)                                                            & 2.76 (0.09)                                                             \\
DrNAS \citep{chen_drnas_2021}        & 4.47 (0.28)                                                            & 2.46 (0.03)                                                             \\
PC-DARTS \citep{xu_pc-darts_2019}    & 4.31 (0.03)                                                                      & 2.57 (0.07)                                                             \\
PDARTS \citep{chen_progressive_2019} & 4.47 (0.15)                                                            & 2.50                                                                    \\
iDARTS \citep{zhang_idarts_2021}     & 4.35 (0.09)                                                            & 2.37 (0.03)                                                             \\ \midrule
TSE-DARTS                                              & 4.64 (0.22)                                                            & 2.73 (0.12)                                                             \\
TSE-DrNAS                                              & 4.35 (0.19)                                                            & 2.90 (0.14)                                                             \\
TSE-PDARTS                                             & 4.61 (0.22)                                                            & 3.19 (0.12)                                                             \\
TSE-DARTS-20                                           & 4.47 (0.18)                                                            & 2.68 (0.04)                                                             \\
DARTS-20 (first-order)                                  & 5.06 (0.39)                                                            & 2.89 (0.11)                                                             \\ \bottomrule
\end{tabular}

\caption{Summary of performances of TSE variants against baselines in the DARTS search space. TSE-DARTS outperforms the baseline DARTS, but the results from applying it to other state-of-the-art algorithms are mixed. DARTS (1st) and (2nd) refer to first and second-order DARTS, respectively.}
\label{tab:darts_orig_results}
\end{table}
It thus appears that searching without the depth gap can be more effective compared to progressive deepening. DARTS keeps the search supernetwork small primarily because of excessive memory requirements due to the continuous relaxation. However, the full-depth search is computationally feasible when using techniques such as gradient accumulation, which allows to trade-off extra compute time for memory, or partial channel connections as in PC-DARTS \citep{xu_pc-darts_2019}. The TSE-DARTS-20 search only takes around 1 day to complete, which is the same as second-order DARTS, and around 23GB of VRAM when using the original 64 batch size. Note that optimizing TSE has the added advantage of making DARTS less memory-intensive due to not using any second-order gradients, hence making it easier to use higher depth supernetworks. At the same time, searching with full depth supernetworks might even be computationally cheaper overall if it were to improve on the search instability in DARTS, reducing the number of seeds that need to be ran to obtain good results.

\subsection{Bias towards shallow architectures}
\label{sec:darts_shallow}
In Section \ref{sec:darts_depth}, we investigated the effect of the number of layers in the supernetwork considered as a fixed hyperparameter. Now, we investigate the depth of the architectural cells themselves which is determined by the search. \citet{shu_understanding_2020} have shown that DARTS tends to be biased towards shallow cells. This is explained by shallower architectures having smoother loss landscapes, which makes them easier to train. DARTS favors them since the gradient-based search is naturally greedy towards architectures which train fast one step ahead.

Following \citet{shu_understanding_2020}, we measure depth as the length of the longest path from input to output within the cell DAG. Figure \ref{fig:darts_depth} shows that TSE-DARTS tends to select significantly deeper cells than normal DARTS. This is inconsistent with the intuition that DARTS is biased towards shallow architectures because they train faster, since optimizing TSE means that the bias towards architectures which achieve low training loss faster should be even more prominent. We propose an explanation for this paradox in Section \ref{sec:darts_ev}. Additionally, we found the depth of TSE-DARTS-20 architectures to be inbetween DARTS and TSE-DARTS, whereas DARTS-20 again found the shallowest cells, same as normal DARTS. It appears that TSE-DARTS with both 8 and 20 layers maintains a roughly constant total depth of the network since searching with 8 cells finds deep individual cells whereas searching with 20 cells finds shallower cells. 

Next, we consider the selection of skip connections, which have been shown to make the loss landscapes of deep neural networks smoother as deep networks tend to otherwise have very non-smooth loss regions \citep{li_visualizing_2018}. Therefore, TSE-DARTS-20 picking more skip connections can also be interpreted in the context of smoothing the loss landscapes, which only becomes necessary when the search supernetwork has 20 layers. Skip connections might have limited utility with only 8 layers because the network is still shallow, and hence are not selected by TSE-DARTS. Our results would thus suggest that the bias towards shallow architectures observed by \citet{shu_understanding_2020} is more strongly related to the higher loss landscape smoothness of such architectures rather than skip connections making the architecture train faster, which is often believed to be the reason for DARTS's bias towards skip connections \citep{zhou_theory-inspired_2020, chen_progressive_2019}. Otherwise, TSE-DARTS would have even stronger overfitting to skip connections than normal DARTS, which is the opposite of what happens empirically.

\begin{figure}[t]
\centering
\begin{subfigure}{.5\textwidth}
  \centering
  \includegraphics[scale=0.46]{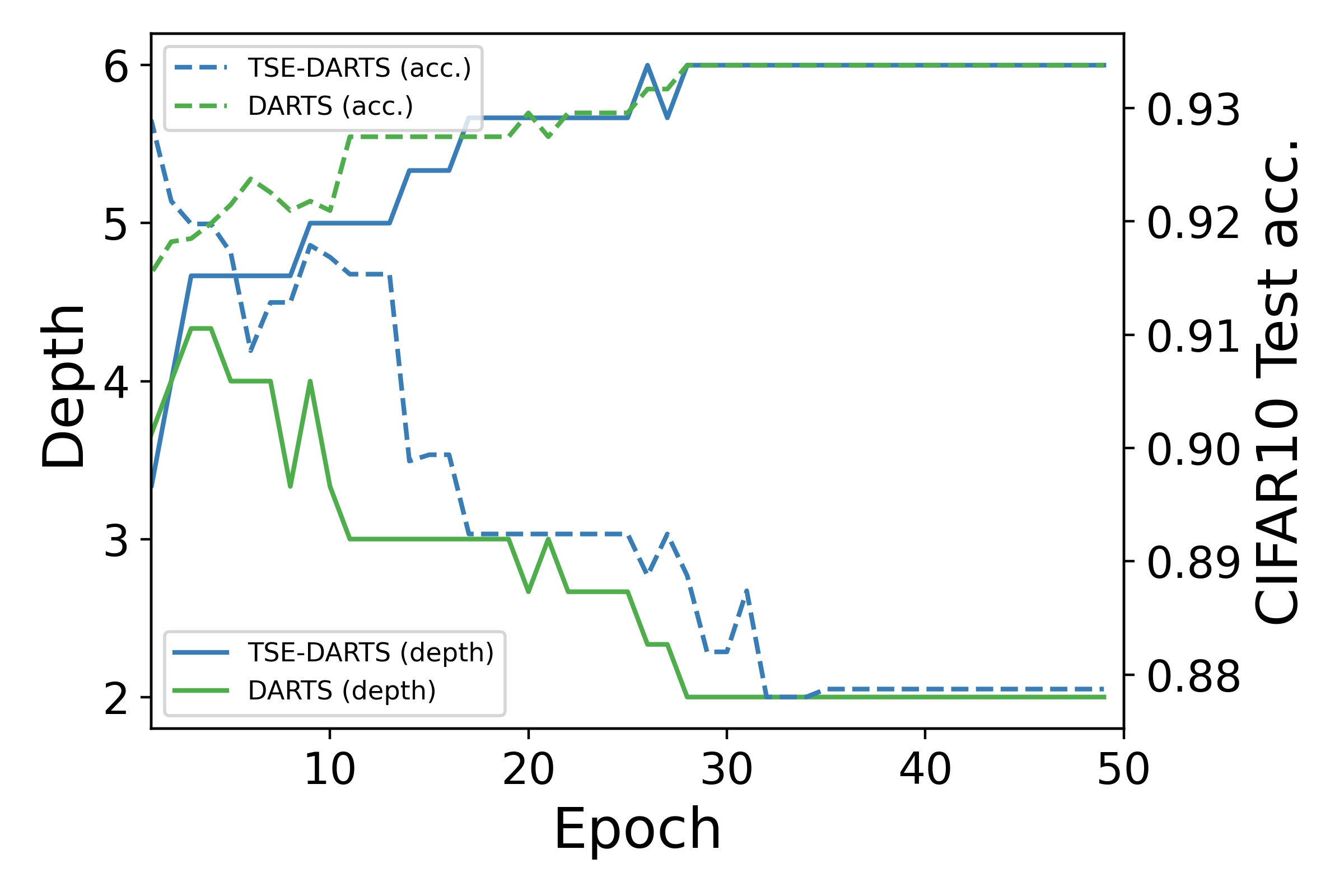}
    \caption{NB101-3 - test acc. and depth}

  \label{fig:nb101_darts_depth_acc}
\end{subfigure}%
\begin{subfigure}{.5\textwidth}
  \centering
  \includegraphics[scale=0.46]{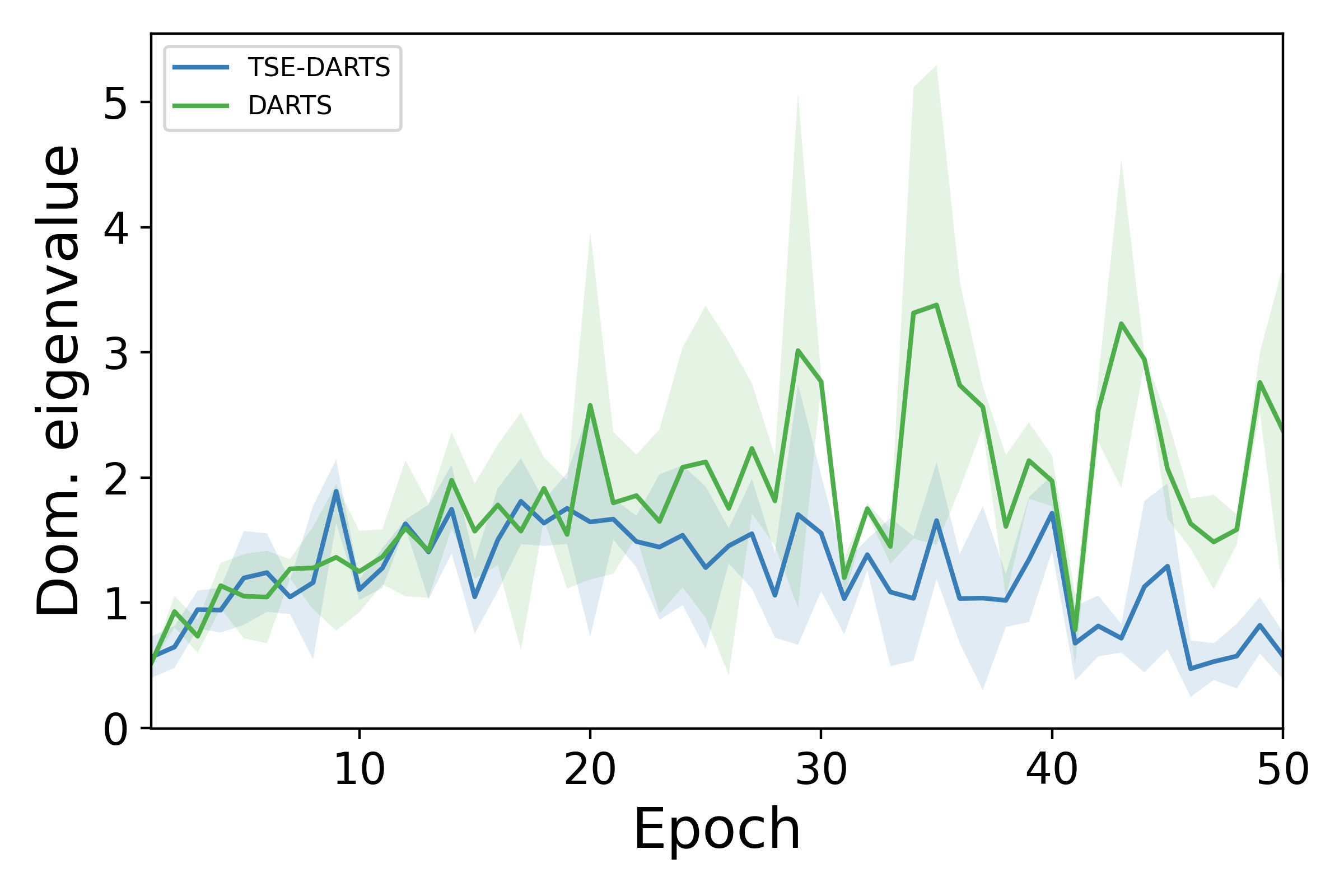}
  \caption{NB101-3 - dom. eigenvalue}
  \label{fig:nb101_darts_depth_conv}
\end{subfigure}
\caption{a) TSE-DARTS only gets worse over time on NB101-3 whereas baseline DARTS achieves top tier performance. Note the strong negative correlation between depth and final performance because both algorithms select $3 \times 3$ convolutions for most if not all operations during the search. In b), we expand the results of \citet{zela_understanding_2019} as discussed in Section \ref{sec:darts_ev}, and show that TSE-DARTS has decreasing eigenvalues along with decreasing performance at the same time.}
\label{fig:nb101_darts_depth}
\end{figure}

We further show that the poor results of TSE-DARTS on NASBench-1shot1 are caused by finding architectures with excessive depth. Both DARTS and TSE-DARTS find architectures which are predominantly if not all 3x3 convolutions, but the performance of DARTS is high because it finds the shallowest cells possible. Note that NASBench-1shot1 search spaces contain no skip connections, which makes it impossible for DARTS to overfit to them. Figure \ref{fig:nb101_darts_depth_acc} shows the performance of the architectures selected by DARTS and TSE-DARTS on NB101-3 alongside the cell depth, where lower depth clearly correlates with better performance. 

Overall, DARTS consistently finds shallow architectures even without any skip connections in the search space while TSE-DARTS finds deep cells. Whether this leads to good or bad performance then depends on the search space. The results here also highlight that selecting the cell topology can be a serious issue in differentiable NAS as the poor performance of TSE-DARTS on NASBench-1shot1 is almost entirely caused by the suboptimal cell topology. However, most work on improving DARTS focuses on the skip connection collapse rather than improving the topology selection.
\subsection{Eigenvalues of the architecture Hessian}
\label{sec:darts_ev}
\citet{zela_understanding_2019} argue that the skip connection overfitting in DARTS is related to the growing dominant eigenvalues of the architecture Hessian. We additionally confirm the trend of rising eigenvalues for DARTS on the NASBench search spaces \citep{dong_nas-bench-201_2020-1, zela_nas-bench-1shot1_2020}, which have been released after the original work by \citet{zela_understanding_2019}. On the other hand, we also show that the behavior of TSE-DARTS eigenvalues is much more varied.

\begin{figure}[t]
\centering
\begin{subfigure}{.5\textwidth}
  \centering
  \includegraphics[scale=0.46]{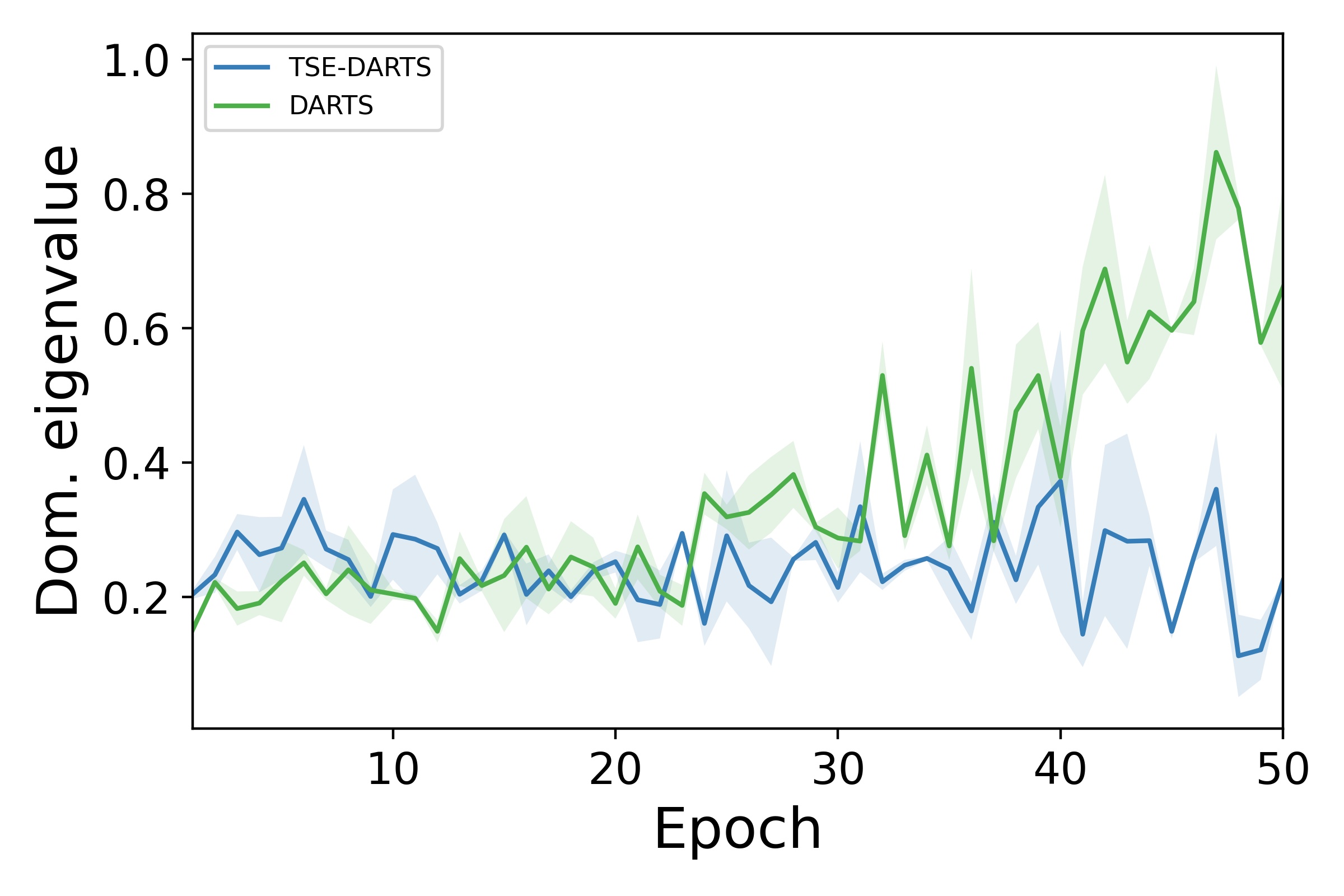}
    \caption{S2 space - dom. eigenvalue}

  \label{fig:darts_s2_ev_acc}
\end{subfigure}%
\begin{subfigure}{.5\textwidth}
  \centering
  \includegraphics[scale=0.46]{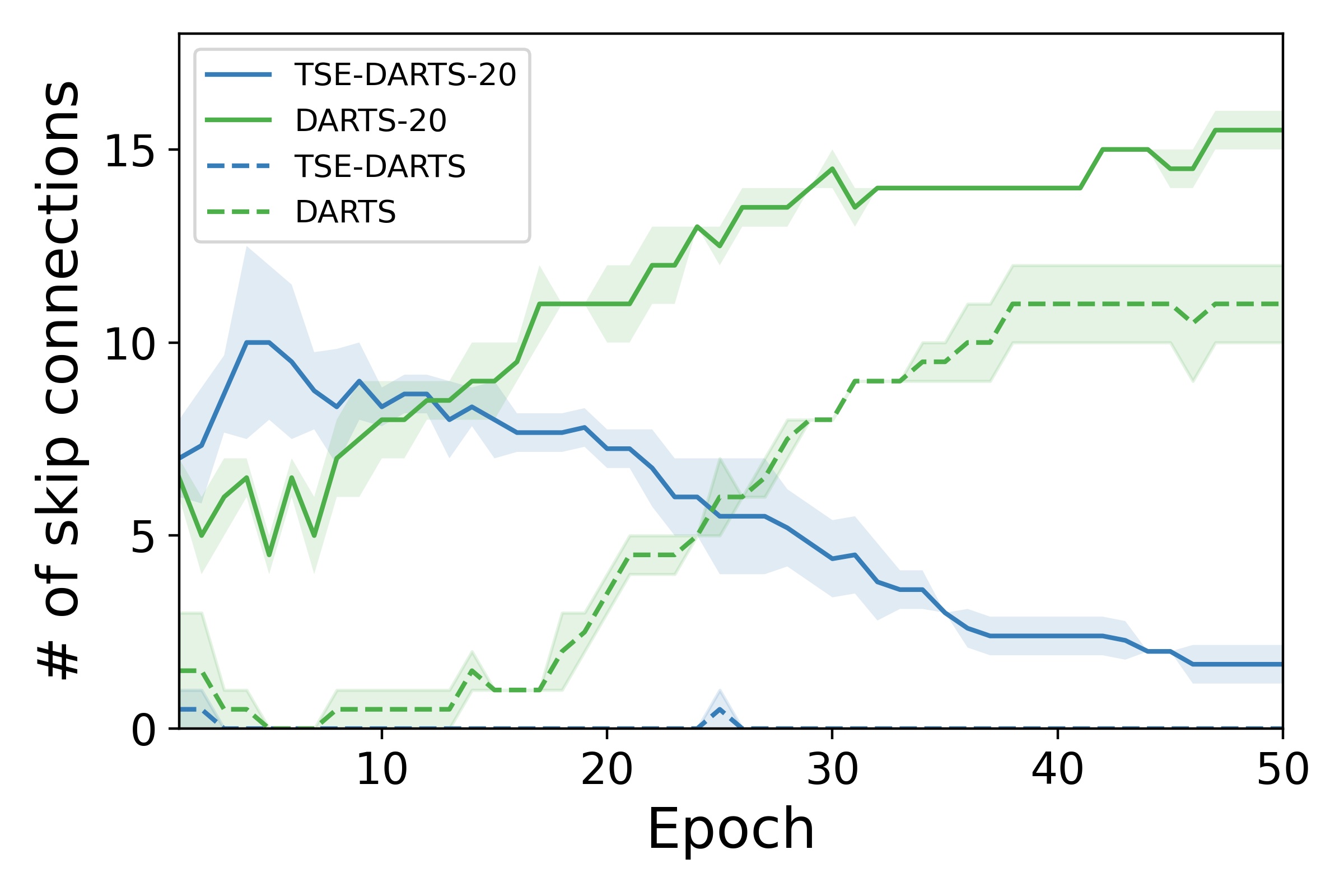}
  \caption{S2 space - \# of skip}
  \label{fig:darts_ev_nb101_ev}
\end{subfigure}

\caption{a) The architecture eigenvalues of TSE-DARTS on the $\text{S2}$ search space do not increase unlike in DARTS. b) shows that the increased search supernetwork depth further exacerbates the overfitting to skip connections for DARTS-20, which picks up to 16 out of 16 operations as skip connections. TSE-DARTS-20 is able to avoid it, and picks a moderate amount of skip connections in the final architecture while shallow TSE-DARTS chooses zero skip connections at all times.}
\label{fig:darts_s2_ev_skip}
\end{figure}

We first reproduce experiments on the reduced $\text{S2}$ search space proposed by \citet{zela_understanding_2019}, which contains only skip connections and $3 \times 3$ separable convolutions for two operations total. Figure \ref{fig:darts_s2_ev_skip} shows that TSE-DARTS picks zero skip connections even on this space whereas DARTS severely overfits to skip connections. DARTS-20 overfits to skip connections even more while TSE-DARTS-20 chooses a small but nonzero number of skip connections, which attests to skip connections being more prominent when the search supernetwork is deep. The architecture eigenvalues increase for DARTS but stay at low levels for TSE-DARTS. Figure \ref{fig:nb101_darts_depth} shows the same experiment on NB101-3, where we see that the eigenvalues for TSE-DARTS actually go down but so does its performance, whereas DARTS has rising eigenvalues and rising performance. On NASBench-201, the eigenvalues increase for both DARTS and TSE-DARTS. However, the performance keeps improving for TSE-DARTS but it gets worse over time for DARTS. We visualize the NASBench-201 results in Appendix \ref{app:nb201_ev}.

Overall, it appears that the eigenvalue phenomenon observed by \citet{zela_understanding_2019} also has significant differences for DARTS and TSE-DARTS. While DARTS always has rising eigenvalues and simultaneously overfits to skip connections, neither appears to be the case for TSE-DARTS. A potential explanation might be that TSE-DARTS is more biased towards architectures with smooth loss landscapes as proposed by \citet{shu_understanding_2020}, which by definition results in smaller architecture Hessian eigenvalues by the virtue of the search algorithm finding architectures without excessively sharp loss regions. This would be in disagreement with the observation in Section \ref{sec:darts_shallow} showing that TSE-DARTS discovers deeper architectures, which should instead have less smooth loss landscapes. Nonetheless, we see that in practice, DARTS has higher eigenvalues despite finding shallower architectures. A plausible explanation might be that the architecture depth is confounded by the architecture selection in DARTS being suboptimal as we had explicitly shown in Section \ref{sec:darts_shallow}. The TSE-DARTS bias towards deep architectures might then be caused by the architecture selection procedure rather than the optimization itself. An explicit treatment of cell topology could therefore allow to better reconcile our results with those of \citet{shu_understanding_2020} and \citet{zela_understanding_2019}. That said, note that analyzing the supernetwork eigenvalues already has the advantage of being independent of architecture selection, and TSE-DARTS having smaller architecture eigenvalues can also be understood as supporting evidence for TSE-DARTS to be finding shallow architectures.

\section{Conclusion}
In this work, we analyzed TSE-DARTS, a variant of DARTS where the architecture is updated to minimize TSE rather than validation loss. Our results show that TSE-DARTS tends to have a stronger performance than DARTS, and we demonstrated that TSE-DARTS also has several interesting qualitative behaviors. In particular, it is free of the skip connection collapse that is known to be a major issue in DARTS. We applied several DARTS diagnostics previously proposed in the literature and shown that TSE-DARTS has fundamentally different biases than the original DARTS. TSE-DARTS appears to be finding significantly deeper architectures than DARTS \citep{shu_understanding_2020}, and its architecture Hessian eigenvalues appear to have different trajectories depending on the search space, whereas the DARTS eigenvalues rise consistently as observed by \citet{zela_understanding_2019}. Our experiments also yielded a concrete example of the depth gap that cannot be fixed with progressive deepening \citep{chen_progressive_2019,chen_drnas_2021}, where TSE-DARTS with an increased number of layers in the search supernetwork starts selecting skip connections despite not picking any otherwise. We have also shown that the DARTS architecture selection via argmax on the architecture coefficients can lead to problems with the cell topology rather than just the selected operations as is commonly assumed.

Ultimately, while TSE-DARTS is able to outperform the baseline DARTS, it appears to have its own distinct set of problems, and it cannot be trivially applied to other state-of-the-art DARTS variants to further advance their performance. However, the follow-up work on DARTS might be effectively overfitted to fixing DARTS's problems, which are no longer present in TSE-DARTS. It is possible that if the same amount of attention was devoted to fixing the problems of TSE-DARTS, it could achieve much better performance. Such extensions are an interesting direction for future work.

\newpage

\begin{ack}
Binxin Ru was supported by the Clarendon Scholarship. Clare Lyle was supported by the Open Philantropy Foundation AI Fellows Program.
\end{ack}


\printbibliography

\section*{Checklist}

\begin{enumerate}

\item For all authors...
\begin{enumerate}
  \item Do the main claims made in the abstract and introduction accurately reflect the paper's contributions and scope?
    \answerYes{}
  \item Did you describe the limitations of your work?
    \answerYes{See the Conclusions section.}
  \item Did you discuss any potential negative societal impacts of your work?
    \answerNo{Our work has no direct societal impact.}
  \item Have you read the ethics review guidelines and ensured that your paper conforms to them?
    \answerYes{}
\end{enumerate}

\item If you are including theoretical results...
\begin{enumerate}
  \item Did you state the full set of assumptions of all theoretical results?
    \answerYes{See Appendix \ref{app:TSE_grad}, which proves our only theoretical result}
	\item Did you include complete proofs of all theoretical results?
    \answerYes{See Appendix \ref{app:TSE_grad}, which proves our only theoretical result.}
\end{enumerate}

\item If you ran experiments...
\begin{enumerate}
  \item Did you include the code, data, and instructions needed to reproduce the main experimental results (either in the supplemental material or as a URL)?
    \answerYes{See supplementary material.}
  \item Did you specify all the training details (e.g., data splits, hyperparameters, how they were chosen)?
    \answerYes{Most hyperparameters are defaults from the corresponding benchmarks. We specified the parameters specific to our method.}
	\item Did you report error bars (e.g., with respect to the random seed after running experiments multiple times)?
    \answerYes{All tabular results include error bars and so do the figures where clarity allows.}
	\item Did you include the total amount of compute and the type of resources used (e.g., type of GPUs, internal cluster, or cloud provider)?
    \answerYes{We specified the search and evaluation costs of our methods as well as the baselines.}
\end{enumerate}

\item If you are using existing assets (e.g., code, data, models) or curating/releasing new assets...
\begin{enumerate}
  \item If your work uses existing assets, did you cite the creators?
    \answerYes{We cite all the NASBench and other relevant baseline code implementations of methods that we adapt. We used no other relevant data or models apart from well-known public datasets such as CIFAR10.}
  \item Did you mention the license of the assets?
    \answerNA{}
  \item Did you include any new assets either in the supplemental material or as a URL?
    \answerNA{}
  \item Did you discuss whether and how consent was obtained from people whose data you're using/curating?
    \answerNA{}
  \item Did you discuss whether the data you are using/curating contains personally identifiable information or offensive content?
    \answerNA{}
\end{enumerate}

\item If you used crowdsourcing or conducted research with human subjects...
\begin{enumerate}
  \item Did you include the full text of instructions given to participants and screenshots, if applicable?
    \answerNA{}
  \item Did you describe any potential participant risks, with links to Institutional Review Board (IRB) approvals, if applicable?
    \answerNA{}
  \item Did you include the estimated hourly wage paid to participants and the total amount spent on participant compensation?
    \answerNA{}
\end{enumerate}

\end{enumerate}


\appendix

\section{Derivation of the exact TSE gradient}
\label{app:TSE_grad}

The derivation here is analogous to those in other work involving differentiation through optimization \citep{lorraine_optimizing_2020, franceschi_forward_2017}. Assume we optimize the weights by SGD, then the weights $w_T$ after $T$ time steps from initial $w_0$ can be written as 
\begin{equation}
\label{eq: w_T}
    w_T = w_0 - \eta \sum_{t=0}^T \nabla_w L_{train}(f(x_t, w_t, \alpha), y_t)
\end{equation}

With this in mind, we will now compute the final training loss gradient after T iterations $\nabla_\alpha L_{train}(f(x_T, w_T, \alpha), y_T)$ as the first step to computing the TSE gradients, where we abbreviate $L_{train}(f(x_T, w_T, \alpha), y_T)$ as $L_{train}^T$:
\begin{equation}
\label{eq: hypergradient}
\begin{split}
    \nabla_{\alpha}L_{train}^T = \frac{\partial L_{train}^T}{\partial \alpha} + \frac{\partial L_{train}^T}{\partial w} \frac{\partial w_T}{\partial \alpha}
\end{split}
\end{equation}
The $\frac{\partial L_{train}^T}{\partial \alpha}$ is sometimes referred to as the direct gradient, and the $\frac{\partial L_{train}^T}{\partial w} \frac{\partial w_T}{\partial \alpha}$ is also known as the indirect gradient or hypergradient. Now given the expression for $w_T$ given in Eq. (\ref{eq: w_T}), we calculate
\begin{equation}
\label{eq:grad_w_T_first_part}
    \begin{split}
        \frac{\partial w_T} {\partial \alpha} & = \frac{\partial}{\partial \alpha} (w_{T-1} - \eta \frac{\partial L_{train}^{T-1}}{\partial w}) \\
        & = \frac{\partial w_{T-1}}{\partial \alpha} - \eta (\frac{\partial^2 L_{train}^{T-1}}{\partial w \partial \alpha} \frac{\partial \alpha}{\partial \alpha} + \frac{\partial^2 L_{train}^{T-1}}{\partial w \partial w} \frac{\partial w_{T-1}}{\partial \alpha}) \\
        & = - \eta \frac{\partial^2 L_{train}^{T-1}}{\partial w \partial \alpha} + (I - \eta \frac{\partial ^2 L_{train}^{T-1}}{\partial w \partial w}) \frac{\partial w_{T-1}}{\partial \alpha}
    \end{split}
\end{equation}
The $\frac{\partial w_{T-1}}{\partial \alpha}$ at the end of Eq. (\ref{eq:grad_w_T_first_part}) gives a recurrent relation that can be further expanded to give 
\begin{equation}
    \begin{split}
        \frac{\partial w_T} {\partial \alpha} & = - \eta \frac{\partial^2 L_{train}^{T-1}}{\partial w \partial \alpha} + (I - \eta \frac{\partial ^2 L_{train}^{T-1}}{\partial w \partial w}) (- \eta \frac{\partial^2 L_{train}^{T-2}}{\partial w \partial \alpha} + (I - \eta \frac{\partial ^2 L_{train}^{T-2}}{\partial w \partial w}) \frac{\partial w_{T-2}}{\partial \alpha}) \\
        & = -\eta \frac{\partial^2 L_{train}^{T-1}}{\partial w \partial \alpha} - \eta (I - \eta \frac{\partial ^2 L_{train}^{T-1}}{\partial w \partial w}) \frac{\partial^2 L_{train}^{T-2}}{\partial w \partial \alpha} + \prod_{0 \leq k < 2}[I-\eta \frac{\partial^2 L_{train}^{T-k-1}}{\partial w \partial w}] \frac{\partial w_{T-2}}{\partial \alpha}
    \end{split}
\end{equation}
If we unroll the whole history until $w_0$, for which it holds $\frac{\partial w_0}{\partial \alpha} = \mathbf{0}$, we get a string of summands that can be summarized as
\begin{equation}
    \frac{\partial w_T} {\partial \alpha} = - \eta \sum_{0 \leq j \leq T}([\prod_{0 \leq k < j} I - \eta \frac{\partial^2 L_{train} ^{T-k-1}}{\partial w \partial w}] \frac{\partial^2L_{train}^{T-j-1}}{\partial w \partial \alpha})
\end{equation}
In total, we have that Eq. (\ref{eq: hypergradient}) is equivalent to

\begin{equation}
\label{eq:single_hypergradient}
    \nabla_{\alpha}L_{train}^T = \frac{\partial L_{train}^T}{\partial \alpha} + \frac{\partial L_{train}^T}{\partial w} (- \eta \sum_{0 \leq j \leq T}([\prod_{0 \leq k < j} I - \eta \frac{\partial^2 L_{train} ^{T-k-1}}{\partial w \partial w}] \frac{\partial^2L_{train}^{T-j-1}}{\partial w \partial \alpha}))
\end{equation}

This is the exact unrolled differentiation hypergradient using the last training loss. The TSE gradient computed over T steps of training is simply equal to a sum of the individual training loss gradients: 
\begin{equation}
    \nabla_{\alpha} TSE = \sum_{t=0}^T (\frac{\partial L_{train}^t}{\partial \alpha} + \frac{\partial L_{train}^t}{\partial w} (- \eta \sum_{0 \leq j \leq t}([\prod_{0 \leq k < j} I - \eta \frac{\partial^2 L_{train} ^{t-k-1}}{\partial w \partial w}] \frac{\partial^2L_{train}^{t-j-1}}{\partial w \partial \alpha})))
\end{equation}

\section{Additional results on NASBench-201 and NASBench-1shot1}
\label{app:extra_nb201_nb101}
We also show the results of TSE-DARTS on all the three datasets in NASBench-201 \citep{dong_nas-bench-201_2020-1} and all the three search spaces in NASBench-1shot1 \citep{zela_nas-bench-1shot1_2020}. For NASBench-201, we additionaly show that training for 150 rather than 50 epochs still retains strong performance and the robustness to skip connection collapse holds even over longer training.

\begin{table}[h!]
\resizebox{\columnwidth}{!}{%

\begin{tabular}{@{}llcc@{}}
\toprule
                        & CIFAR10                          & CIFAR100     & ImageNet16-120 \\ \midrule
REA   \citep{real_regularized_2019}                  & 94.02 (0.31)                     & 72.23 (0.95) & 45.77 (0.80)   \\
Random Search  \citep{bergstra_random_2012}        & 93.90 (0.26)                     & 71.86 (0.89) & 45.28 (0.97)   \\
SPOS        \citep{li_random_2020}            & 91.05 (0.66)                     & 68.26 (0.96) & 40.69 (0.36)   \\
GDAS  \citep{dong_searching_2019}                  & \multicolumn{1}{c}{93.23 (0.58)} & 68.17 (2.50) & 39.40 (0.00)   \\
ENAS      \citep{pham_efficient_2018}              & 93.76 (0.00)                     & 70.67 (0.62) & 41.44 (0.00)   \\
DrNAS       \citep{chen_drnas_2021}            & 93.76 (0.00)                     & 68.82 (2.06) & 41.44 (0.00)   \\
DARTS (first-order) \citep{liu_darts_2018}     & 59.84 (7.84)                     & 61.26 (4.43) & 37.88 (2.91)   \\
DARTS (second-order) \citep{liu_darts_2018}    & 65.38 (7.84)                     & 60.49 (4.95) & 36.79 (7.59)   \\ \midrule
TSE-DARTS              & 92.66 (0.00)                     & 68.13 (2.18) & 36.60 (0.00)   \\
TSE-DARTS (150 epochs) & 89.89 (2.52)                     & 71.03 (0.66) & 33.75 (0.00)   \\
TSE-DrNAS              & 93.76 (0.00)                     & 71.11 (0.00) & 41.44 (0.00)   \\ \bottomrule
\end{tabular} %
}
\caption{Baseline DARTS overfits to all skip connections on NASBench-201, which leads to very poor performance. TSE-DARTS prevents this and achieves $92.66\%$ test set accuracy. Most baselines were reprinted from \citet{dong_nats-bench_2021} except the DrNAS results, which are our own. Notably, we were unable to reproduce the original results from \citet{chen_drnas_2021} with near-perfect performance using the public code.}
\label{tab:nb201_darts_full}
\end{table}

\begin{table}[h!]
\centering
\begin{tabular}{@{}llcc@{}}
\toprule
          & NB101-1      & NB101-2      & NB101-3      \\ \midrule
DARTS     & 93.33 (0.01) & 93.37 (0.00) & 93.35 (0.01) \\
TSE-DARTS & 91.86 (0.68) & 90.15 (0.00) & 87.87 (0.00) \\ \bottomrule
\end{tabular}
\caption{On NASBench-1shot1, the performance of TSE-DARTS is quite weak overall and significantly trails baseline DARTS on all the three search spaces. NB101-1, NB101-2 and NB101-3 abbreviate the NASBench-1shot1 search spaces 1, 2 and 3, respectively.}
\label{tab:nb101_darts}
\end{table}

\newpage

\section{Best discovered architectures by TSE-DARTS}
\label{app:best_arch}
\begin{figure}[h!]
\centering

\begin{subfigure}{.49\textwidth}
  \centering
  \includegraphics[width=\linewidth]{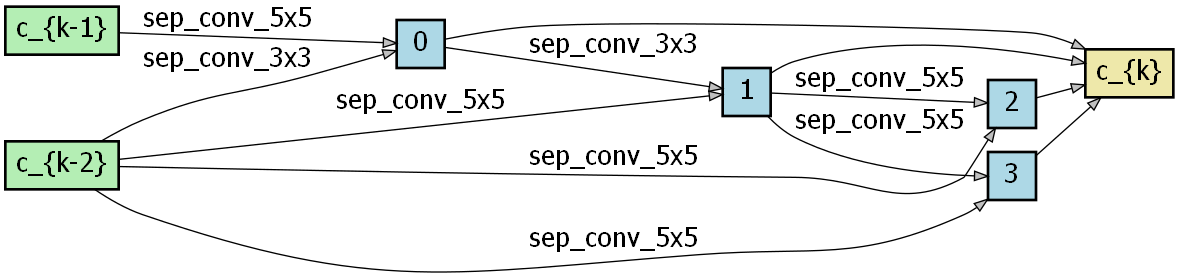}
  \caption{TSE-DARTS-8 - normal cell}
  \label{fig:darts_TSE8_cell}
\end{subfigure}
\begin{subfigure}{.49\textwidth}
  \centering

  \includegraphics[width=\linewidth,height=0.15\textheight]{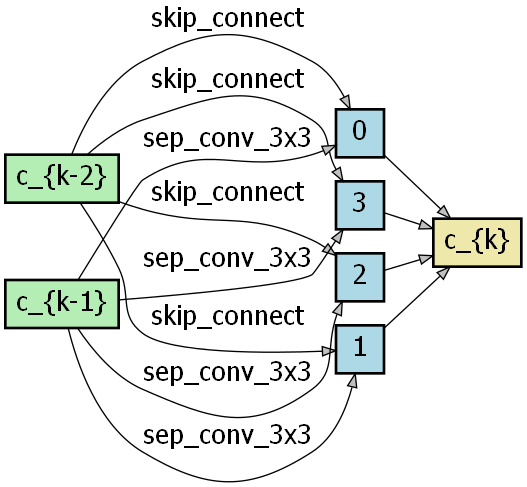}
  \caption{TSE-DARTS-20 - normal cell}
  \label{fig:darts_TSE20_cell}
\end{subfigure}

\medskip

\begin{subfigure}{.49\textwidth}
  \centering
  \includegraphics[width=\linewidth]{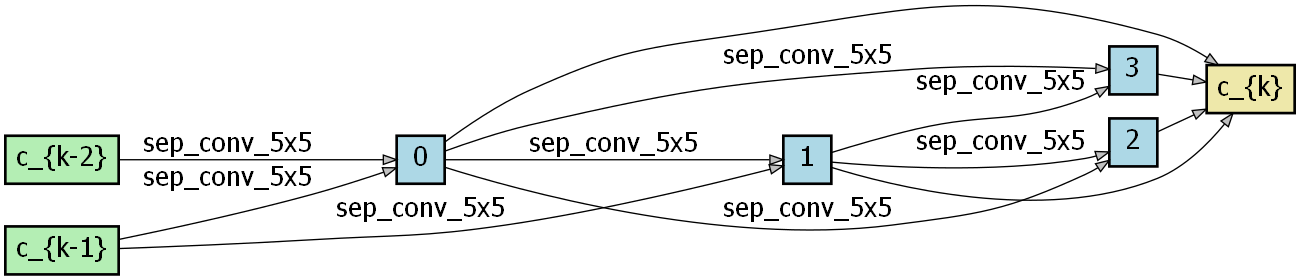}
  \caption{TSE-DARTS-8 - reduction cell}
  \label{fig:darts_TSE8_cell_reduction}
\end{subfigure}
\begin{subfigure}{.49\textwidth}
  \centering

  \includegraphics[width=\linewidth,height=0.15\textheight]{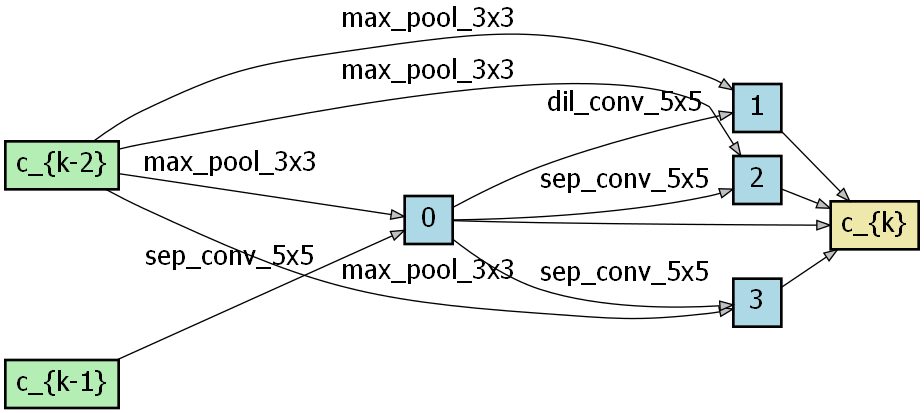}
  \caption{TSE-DARTS-20 - reduction cell}
  \label{fig:darts_TSE20_cell_reduction}
\end{subfigure}

\caption{Overview of the best normal and reduction cells found by TSE-DARTS-8 and TSE-DARTS-20 on the DARTS search space.}
\label{fig:darts_cell_examples}
\end{figure}

Figure \ref{fig:darts_cell_examples} shows the final best architectures found by TSE-DARTS and TSE-DARTS-20. Notably, TSE-DARTS finds significantly deeper architectures for both the normal and reduction cells than TSE-DARTS-20.

\newpage

\section{TSE-DARTS implementation}

\label{app:TSE_darts_implementation}

We showcase an example PyTorch implementation of Algorithm \ref{algo:TSE_darts} (TSE-DARTS) to demonstrate that TSE is very easy to implement on top of standard training loops. The most important part is on lines 12-17, which show that it is trivial to accumulate the approximate TSE gradient by not zeroing out the architecture gradients after each weight update. All implementations of TSE-DARTS variants proceed analogously to this example.
\begin{minted}[fontsize=\scriptsize, linenos]{python}
def train_TSE(train_queue, network, criterion, w_optimizer, a_optimizer, T=100):
    train_iter = iter(train_queue)
    network.train()      

    for unrolling_step in range(math.ceil(len(train_queue)/T)):
      # format_input_data outputs a list of T (input, output) pairs
      all_base_inputs, all_base_targets = format_input_data(train_iter, T=T) 
      network.zero_grad()
      model_init = deepcopy(network.state_dict()) # Save weights before unrolling so they can be restored later

      # Step 1 of Algorithm 3 - do the unrolling over 100 steps to collect TSE gradient
      for (base_inputs, base_targets) in zip(all_base_inputs, all_base_targets):
          logits = network(base_inputs)
          base_loss = criterion(logits, base_targets)
          base_loss.backward()
          w_optimizer.step() # Train the weights during unrolling as normal,
          w_optimizer.zero_grad() # but the architecture gradients are not zeroed during the unrolling


      # Step 2 of Algorithm 3 - update the architecture encoding using accumulated gradients
      a_optimizer.step()
      
      a_optimizer.zero_grad() # Reset to get ready for new unrolling
      w_optimizer.zero_grad()
      
      new_arch_params = deepcopy(network.arch_params) # Temporary backup for new architecture encoding
      
      network.load_state_dict(model_init) # Old weights are loaded, which also reverts the architecture encoding 
      for p1, p2 in zip (network.arch_params, new_arch_params):
        p1.data = p2.data
        
      # Step 3 of Algorithm 3 - training weights after updating the architecture encoding
      for (base_inputs, base_targets) in zip(all_base_inputs, all_base_targets):
        logits = network(base_inputs)
        base_loss = criterion(logits, base_targets)
        base_loss.backward()
        w_optimizer.step()
        
        w_optimizer.zero_grad()
        a_optimizer.zero_grad()


\end{minted}
\newpage

\section{Eigenvalues on NASBench-201}
\label{app:nb201_ev}

\begin{figure}[h!]
\centering
\begin{subfigure}{.5\textwidth}
  \centering
  \includegraphics[scale=0.46]{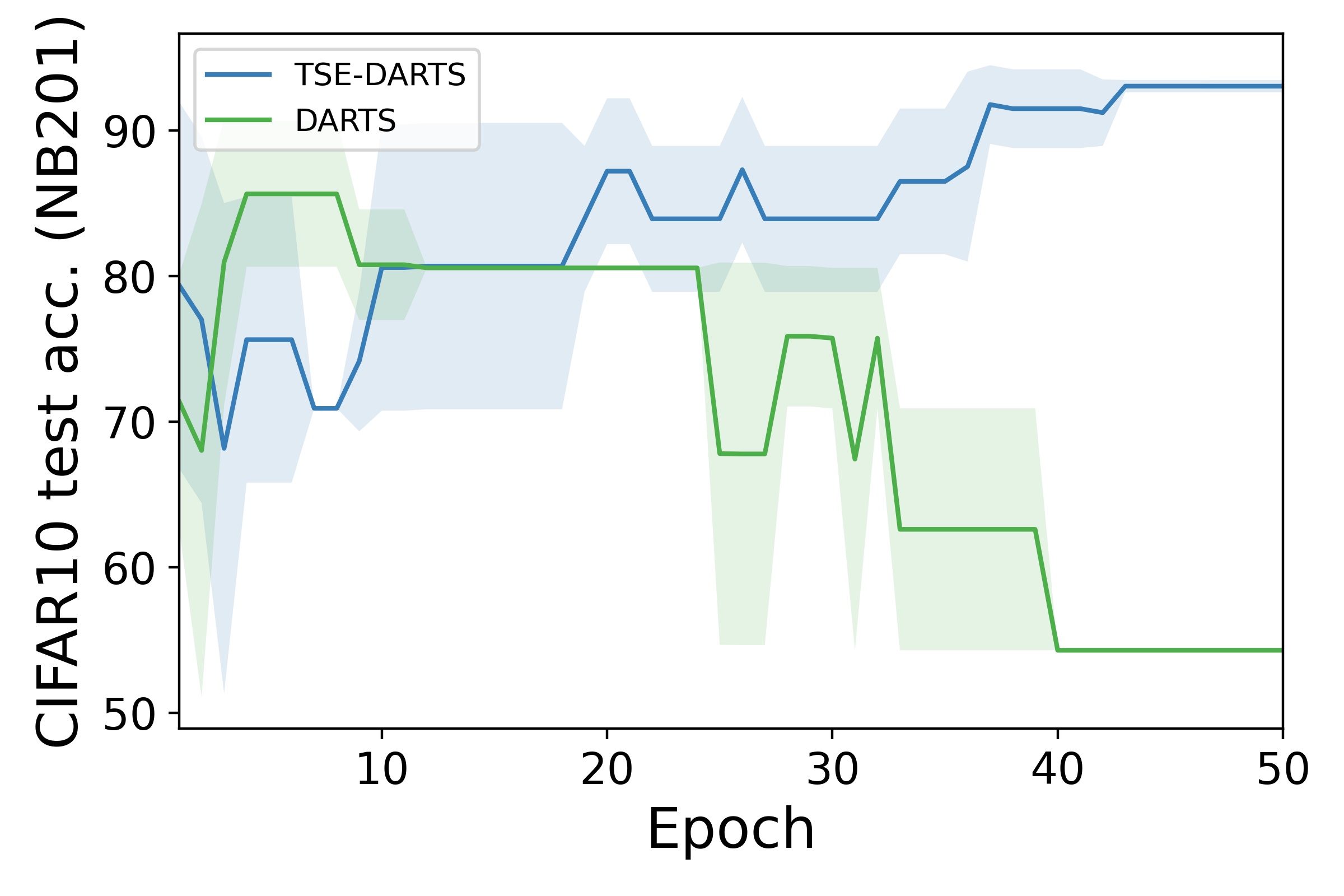}
    \caption{NB201 - test acc. on CIFAR10}

  \label{fig:nb201_ev_acc}
\end{subfigure}%
\begin{subfigure}{.5\textwidth}
  \centering
  \includegraphics[scale=0.46]{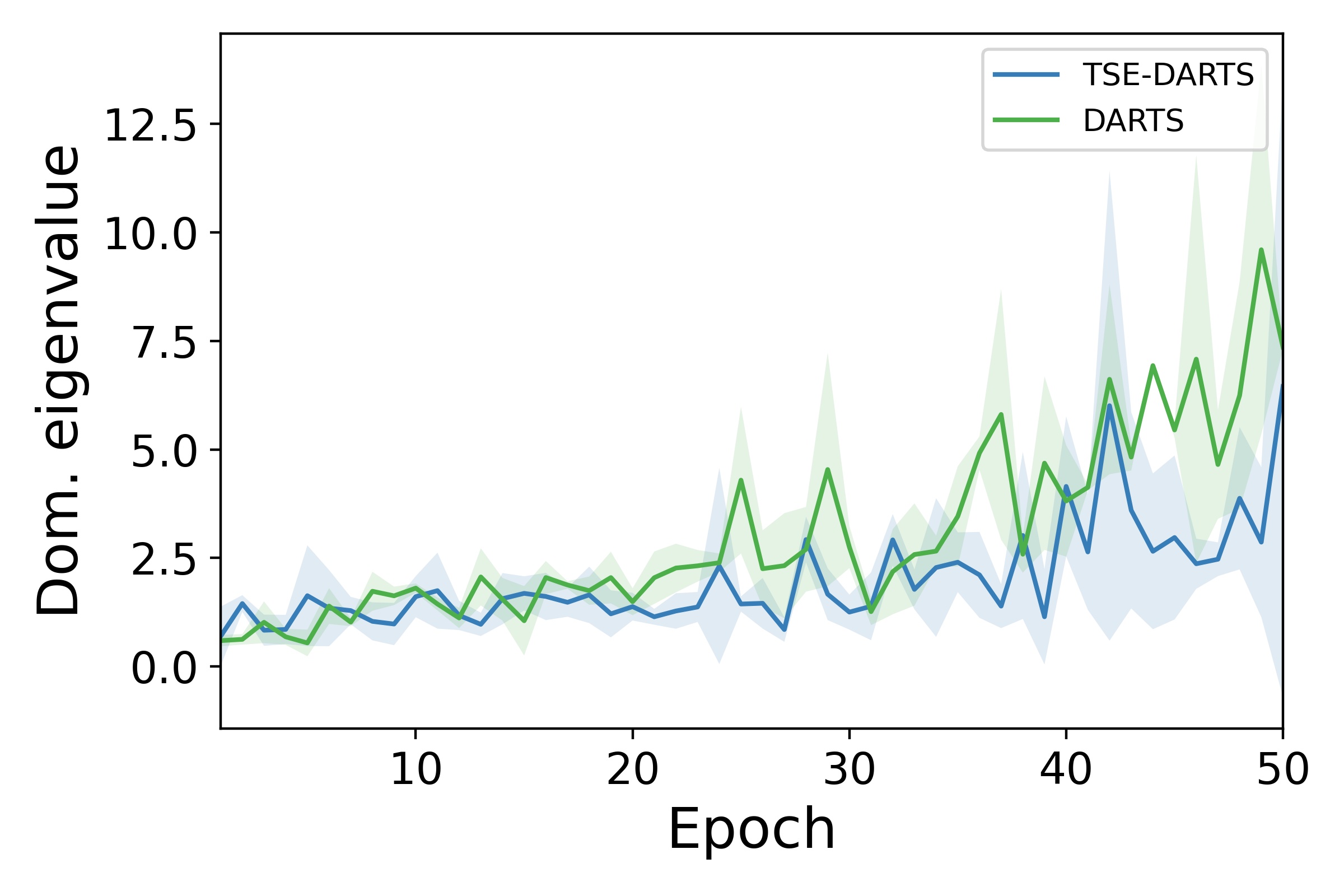}
  \caption{NB201 - dom. eigenvalue}
  \label{fig:nb201_ev_ev}
\end{subfigure}
\caption{On NASBench-201, both DARTS and TSE-DARTS have rising eigenvalues. However, the DARTS performance keeps going down while TSE-DARTS performance increases consistently.}
\label{fig:nb201_ev}
\end{figure}

\end{document}